# Spiking sampling network for image sparse representation and dynamic vision sensor data compression

**Chunming Jiang**[1] **and Yilei Zhang\***[1]

[1]Mechanical engineering, University of Canterbury, New Zealand

\* Correspondence:   Yilei Zhang   Email: yilei.zhang@canterbury.ac.nz

**Abstract**

Sparse representation has attracted great attention because it can greatly save storage re- sources and find representative features of data in a low-dimensional space. As a result, it may be widely applied in engineering domains including feature extraction, compressed sensing, signal denoising, picture clustering, and dictionary learning, just to name a few. In this paper, we propose a spiking sampling network. This network is composed of spiking neurons, and it can dynamically decide which pixel points should be retained and which ones need to be masked according to the input. Our experiments demonstrate that this approach enables better sparse representation of the original image and facilitates image reconstruction compared to random sampling. We thus use this approach for compressing massive data from the dynamic vision sensor, which greatly reduces the storage requirements for event data.

## 1.    Introduction

Sparse signal representation has been demonstrated to be a highly effective technique for obtaining, representing, and compressing high-dimensional signals. Important signal classes, such as audio and images, have sparse representations with respect to a particular basis (e.g., Fourier and wavelet bases) or the concatenation of them. Furthermore, efficient and demonstrably successful techniques based on convex optimization or greedy pursuit are available for computing such high-fidelity representations [1].

Sparse representation is not only widely used in signal processing but is also useful for vision tasks. In the past few years, sparse representation has been applied in face recognition [2-8], image super-resolution [9], motion and data segmentation [10], denoising and painting [11-13], background modeling [14, 15], photometric stereo [16], and image classification [17, 18]. In almost all these applications, the use of sparse representation has achieved impressive results.

The capacity of sparse representations to reveal semantic information is influenced in part by a simple but crucial attribute of the data: despite the images' (or their features') naturally high dimensionality, images belonging to the same class demonstrate degenerate structure in many applications. In other words, they are situated on or close to low-dimensional subspaces, submanifolds, or stratifications. If a collection of representative samples is obtained for this low-dimensional distribution, we could anticipate that a typical sample will have a sparse (potentially learnt) representation over this basis. If appropriately computed, such a sparse representation might naturally encode semantic visual information [19].

Deep learning has developed rapidly in recent years, and kinds of artificial neural networks [20-

23] have been applied to a wide variety of tasks. In deep learning, many works have attempted to introduce sparse coding into neural networks. They usually mask certain input information randomly, which can be considered as a certain kind of sparse coding. BERT and GPT, for instance, are very effective pre-training techniques for NLP. In order to train models to anticipate the missing information, these techniques hold out a piece of the input sequence. There are tons of evidence that these techniques generalize very well and that the pre-trained representations perform admirably across a wide range of downstream tasks.

Methods exist for encoding masked images pick up representations from masked images that have been distorted. Convolutional networks are used by the Context Encoder [24] to fill in significant missing sections. iGPT [25] guesses unknown pixels based on pixel sequences. In the ViT study [26], masked patch prediction for unsupervised learning is considered. The most recent technique for predicting discrete tokens arises from BEiT. MAE [27] tries different masking methods to train the autoencoder which can be adopted to serve as the pre-training model. In most cases, the mask is a randomly generated sampling matrix. Recovering signals from fewer data gathered by a random measurement matrix is efficient. However, they constantly have issues with unclear quality of reconstruction [28].

Spiking neural networks (SNNs) are receiving increasing attention due to their low power consumption and bio-plausibility. Neurons in SNNs receive spike trains that either increase or decrease their membrane potential over time. When the membrane potential exceeds a certain threshold, the neuron fires one spike to next layer's neurons and reset its potential. These characteristics are similar to the way the brain transmits and processes information. It is therefore regarded as the next-generation neural network [29].

Since the spiking neural network (SNN) naturally outputs only 0 and 1 state values, we shall design a spiking autoencoder that generates a binary mask based on the input, where 0 means a certain pixel is not sampled and 1 means that the pixel is sampled at the input stage. Such a mask is multiplied with the input image to obtain the sampled image.

In this paper, we propose a novel sampling network based on spiking neural networks, which is able to dynamically sample the input images, retain the valid pixels and remove the redundant pixels to output a sparse representation of the inputs. We validate its advantages over random sampling for network reconstruction on MNIST and CIFAR-10 datasets. Besides, we apply it to the compression of data generated by event cameras, which greatly reduces the space needed for data storage.

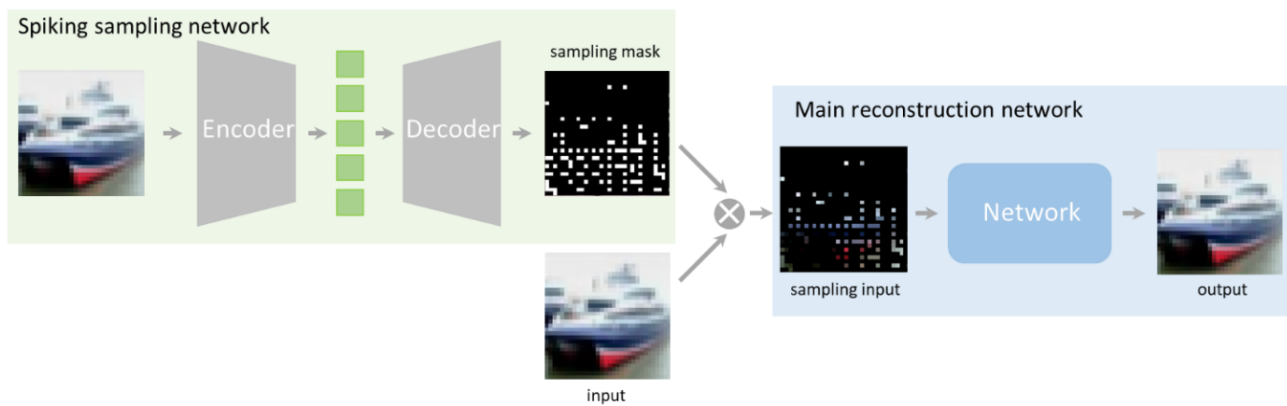

Fig.1 Architecture of the spiking sampling network. The output of spiking sampling network is a mask of the same size as the input.

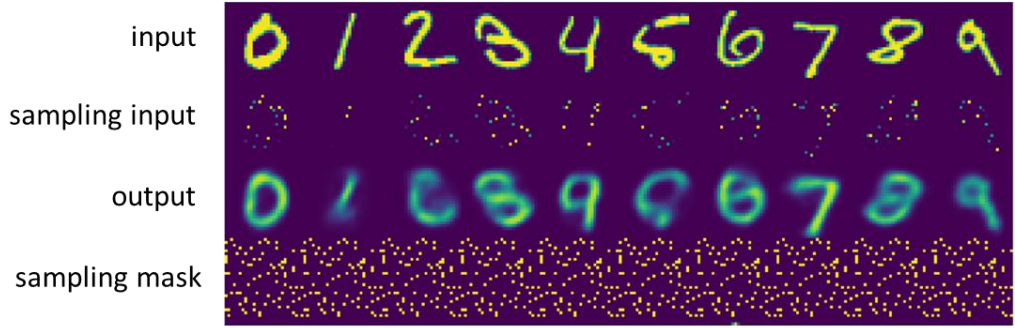
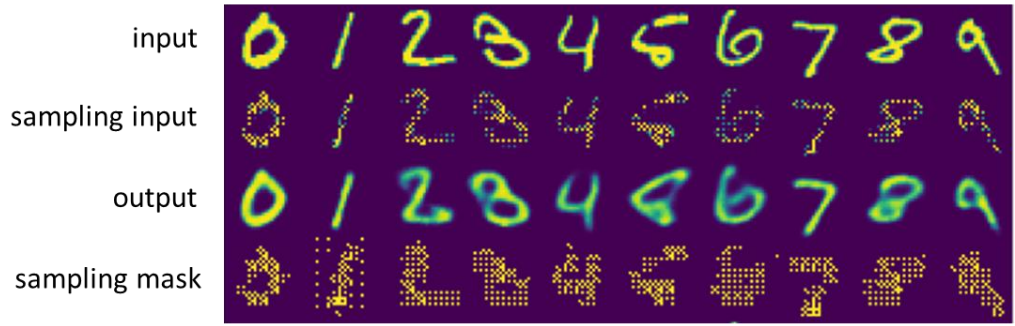

(a)

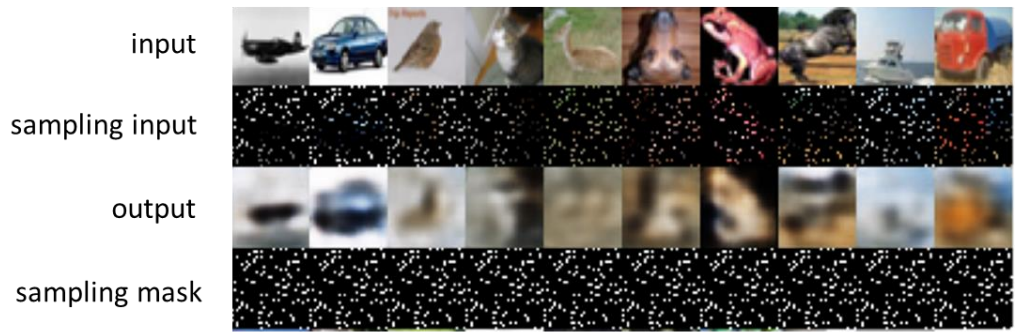
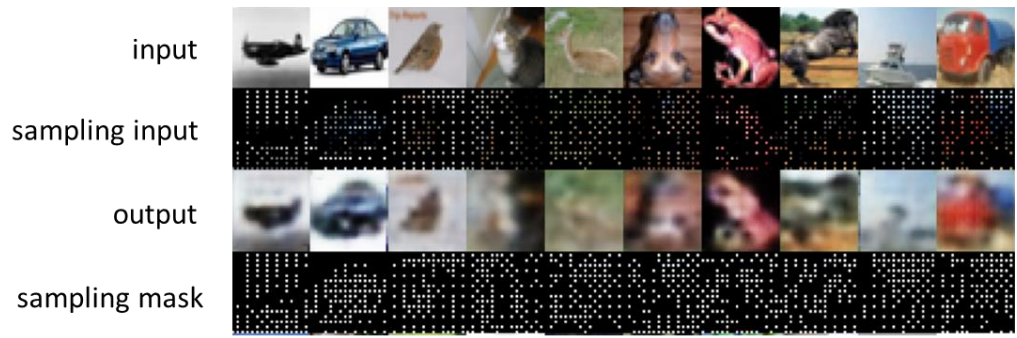

(b)

Fig.2 Comparison between random sampling and spiking sampling on (a) MNIST and (b) CIFAR-10.

## 2. Methods

## 2.1. Leaky Integrate-and-Fire (LIF) model

SNNs adopt spike trains as information carriers between neurons. Every spiking neuron in a SNN receives and emits spikes. The LIF neuron model is a popular bio-inspired simplified model for describing the dynamics of spiking neurons. The dynamics of the LIF model are defined by [26]

$$H(t) = \lambda * V(t-1) + \sum_i w_i x_i(t) \tag{1}$$

$$S(t) = \begin{cases} 1, & H(t) \geq V_{th} \\ 0, & H(t) < V_{th} \end{cases} \tag{2}$$

$$V(t) = H(t)(1 - S(t)) + V_{reset} * S(t) \tag{3}$$

where $H(t)$ and $V(t)$ represent the membrane potentials before and after firing a spike at time $t$, respectively. $V_{th}$ denotes the firing threshold, which is 1 in this paper. $V_{reset}$ is the resting potential which is 0. $S(t)$ denotes the output of a neuron at time $t$, $w_i x_i(t)$ is the $i$-th weighted pre-synaptic input at time $t$, and $\lambda$ is the decaying time constant with a value of 0.5.

## 2.2. Architecture and training of spiking sampling network

Figure 1 illustrates the architecture of the spiking sampling network. It is actually an autoencoder composed of spiking neurons. The neurons of all layers except the last layer have predefined thresholds $V_{th}$ = 1. In the last layer, the threshold of the neurons is not a fixed value but varies dynamically with the input. The spiking neurons in the last layer only accumulate potentials over time and do not fire spikes until the last time step $T$. At instant $T$, we rank the accumulated potentials of all neurons from largest to smallest, and if we need to sample $N$ pixel points, the Nth largest potential is used as the threshold $V_{th}$ so that the number of neurons that fire spikes is $N$.

The output of the spiking sampling network is a sampling mask that has the same size as the input. This sampling mask will be multiplied by the actual input image to preserve the selected pixels. These pixels will be used as input to the main reconstruction network that is used for reconstructing the original image.

In contrast to the commonly used random sampling, our sampling scheme is implemented by a spiking neural network whose parameters can be optimized via back propagation. Thus, the network is able to automatically sample different pixel points for different inputs, depending on the main vision task.

## 2.3. Data compression of dynamic vision sensor

Dynamic Vision Sensor (DVS), also called the event camera, is based on the principle of biosensing, which means that they report only the ON/OFF triggering of luminance in the observed scene [30]. Unlike conventional RGB cameras, which acquire raw data in a two-dimensional matrix, in event cameras, each pixel works independently and asynchronously, reporting changes in luminance as they happen or remain inactive while light intensity is constant [31]. In real-time interaction systems like robotics, drones, and autonomous driving, the DVS's distinctive features provide benefits over traditional vision sensors. In the near future, cloud and edge computing will be used to execute the majority of the services that do object/gesture recognition or classification.

Therefore, in order to interpret visual data, these services would need to send spike events to cloud or edge servers [32]. Real-time transmission is also necessary in many circumstance. Despite the inherent compression offered by the neuromorphic sensing technology, further compression of the generated data may be advantageous for sending such data over Internet of Things (IoT), Internet of Things (IoV), and Industrial IoT (IIoT) situations [30]. Since the data storage and transmission bandwidth for onboard DVS processing and transmission are both limited, the compression of neuromorphic spikes is still a difficult problem that needs quick solves.

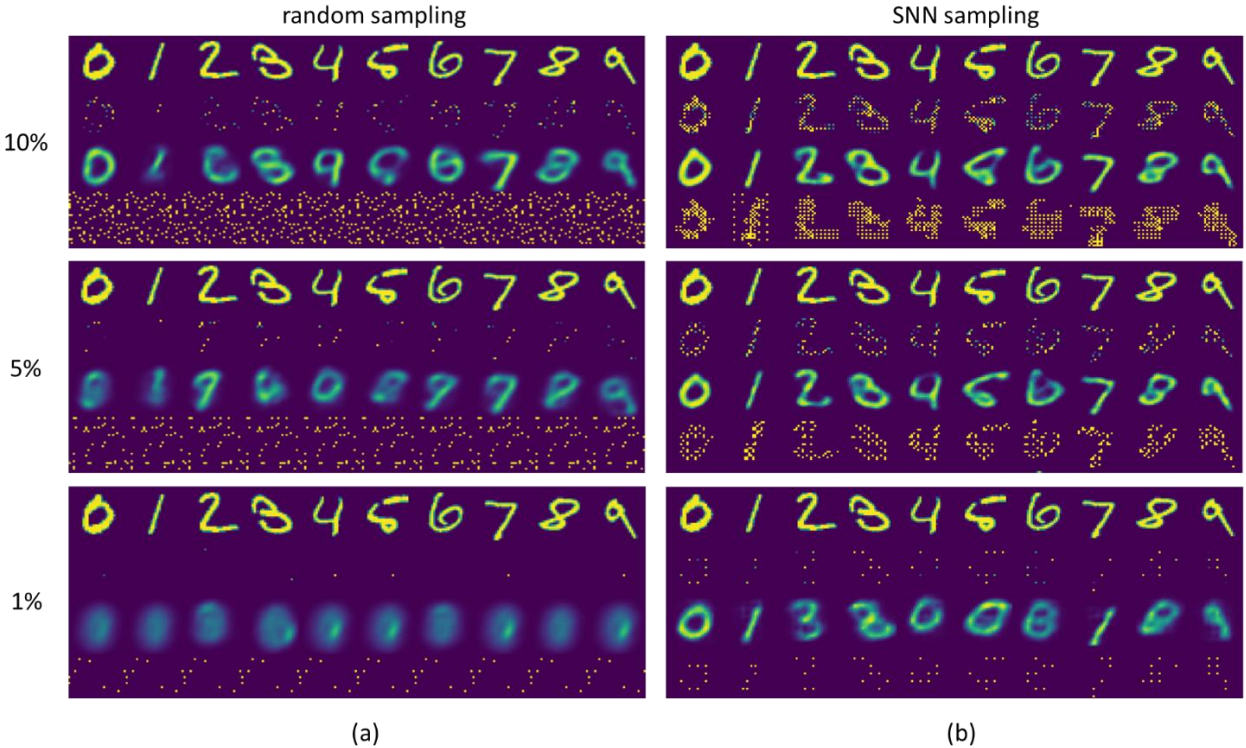

Fig. 3 Different sampling rate comparison of random sampling and spiking sampling on MNIST

## 3. Experiments

### 3.1. Image reconstruction comparison

In Figure 1, we compare the effect of random and spiking sampling on image reconstruction at a sampling rate of 10%. We conduct the experiments on MNIST and CIFAT-10 datasets. The details of hyper-parameter selection and network architectures are listed in the appendix. We can see that for random sampling, the sampling positions are uniformly distributed over the entire image. The spiking sampling, on the other hand, changes the sampling positions, depending on the input. For the MNIST dataset, the spiking sampling focuses on sampling over the figures while ignoring the surrounding background, and for the CIFAR-10 dataset, the sampling density is relatively small in the parts of the image with clean areas and increases in the areas with complex texture. For MNIST, spiking sampling reconstructs images more clearly than that of random sampling. For CIFAR-10, the color and shape of the reconstructed images using the pulse sampling network are more accurate than those from random sampling. Consequently, the pixels sampled by the spiking sampling are significantly more conducive to image reconstruction. This shows that the spiking sampling can effectively make the sparse representation of images.

Figure 3 shows the difference between reconstructed images with random sampling and SNN sampling at different sampling rates, respectively. It can be seen that random sampling at 10% sampling rate can no longer correctly distinguish all the reconstructed digits (e.g., digit 4), while using SNN sampling at 5% sampling rate can still clearly reconstruct all the images. At a sampling rate of 1%, random sampling is completely useless, while SNN sampling is still able to reconstruct some of the digits. Even with few sampling points, SNN sampling is still able to distribute the sampling points over the numbers to be reconstructed, effectively providing a sparse representation of the image. This indicates that the SNN network really learns the pixel points that are useful for reconstructing the image.

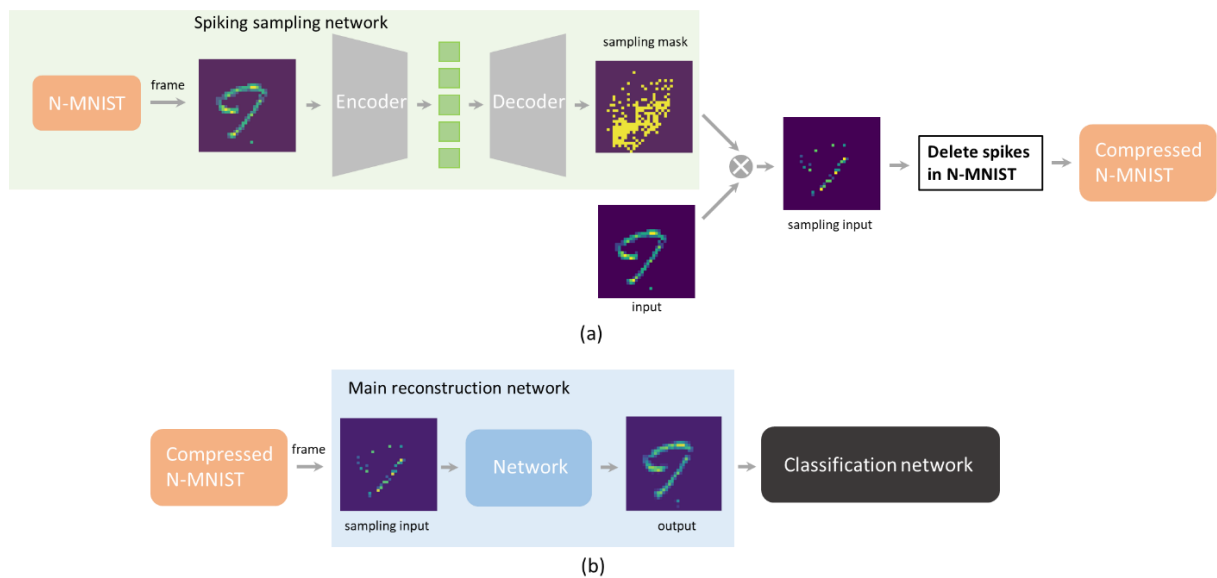

Fig. 4 (a) Compression of N-MNIST dataset. (b) Classification validation of compressed N-MNIST dataset.

### 3.2. Event data compression

A spike event is composed of four basic elements, represented by a tuple {X, Y, t, p}: the spatial addresses X and Y, the timestamp t and the polarity p. The unique spike emission mechanism enables DVS to meet low bandwidth, low power, and low latency requirements. The unique spike emission mechanism enables DVS to meet the requirements of low bandwidth, low power consumption and low latency. At the same time, it also brings a huge amount of data. As an example, the commonly used handwritten numeric dataset MNIST only occupies about 11MB of storage space after compression, while the N-MNIST event dataset generated by MNIST still requires more than 1GB of storage space even after compression. Large datasets often require tens or even hundreds of GB of storage space, which puts a lot of storage pressure.

Since the output of DVS is very different from traditional frame-based image sequences, existing computer vision techniques cannot be directly applied to neuromorphic spike event sequences. Integrating the original event stream into frame data is a common processing method. Therefore, we first render the spatio-temporal coordinates {X, Y, t} and polarity p of the neuromorphic sequences into frames before inputting them to the network. This rendering technique can be referenced in [26].

After training, we keep the spikes corresponding to the sampled pixels and remove the spikes corresponding to the unsampled pixels in the N-MNIST dataset based on the masks generated by the spiking sampling network (see Figure 4(a)).

To verify the validity of our retained event data, we do the classification task on the censored event dataset by a classification network (see Figure 4(b)), and the result is shown in Figure 5(a). It can be seen that when we use SNN sampling, the classification accuracy has only a slight accuracy loss at both 5% and 10% sampling rates, while the event data retained using random sampling causes a large accuracy loss on N-MNIST. Figure 5(b) shows the data size compared to the original N-MNIST dataset when it samples 5% and 10% by random sampling and spiking sampling network, respectively. Since spiking sampling focuses more on the spike-dense region in the image, it retains more spikes than random sampling at the same sampling rate, and the corresponding compression rate is somewhat smaller. Comparing figures 5(a) and 5(b), the data size is reduced by 84% and 88% at a sampling rate of 10% and 5%, respectively, with a slight loss of accuracy, indicating that the spiking sampling network is able to sparsely represent the event dataset effectively.

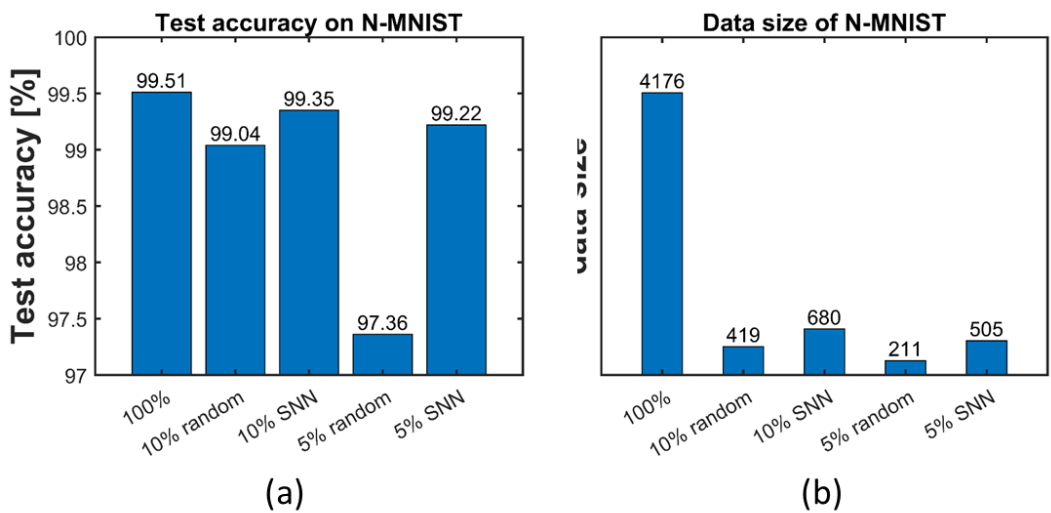

Fig. 5 (a) Test classification accuracy on N-MNIST with different sampling method and rate. (b) data size comparison after different compressing rate by spiking sampling. The numbers on the bars represent the average number of spikes retained per sample for the dataset.

### 3.3. Specificity and universality

In section 2.2, we know that image recovery needs to go through two steps, sampling and reconstruction. From the previous section, it can be verified that sampling method has a great impact on the reconstruction result. The main reconstruction networks obtained by taking different sampling methods for training also differ. In this section, we verify the sensitivity of the main reconstruction network to the sampling methods.

After training, we are able to obtain two reconstruction networks, which target random sampling and spiking sampling reconstruction, respectively. Now we do both kinds of sampling separately and input the sampled pixels to the same reconstruction network to compare the reconstructed results. Figure 6 shows the output difference of main reconstruction network trained by random sampling, when we use random sampling and spiking sampling for test. We can see that even if we use random sampling during training, the quality of the image reconstructed by spiking sampling is no worse than random sampling during test. This shows that the main reconstruction network trained with random sampling has the good universality, and it is less sensitive to different sampling methods. Figure 7 shows the output difference of main reconstruction network trained by spiking sampling, when we use random sampling and spiking sampling for test. Random sampling has a great impact on this main reconstruction network, and the reconstructed images are poor. It demonstrates that the main reconstruction network trained with spiking sampling is more susceptible to the influence of the sampling method and therefore it is more specific to the sampling method. Therefore this sampling method has more potential applications in terms of data privacy and security.

In summary, we conclude that spiking sampling enables higher reconstruction quality, but lead main reconstruction network to be specific with the sampling method; while random sampling makes the reconstruction process more difficult, but make main reconstruction network have better universality on the sampling method.

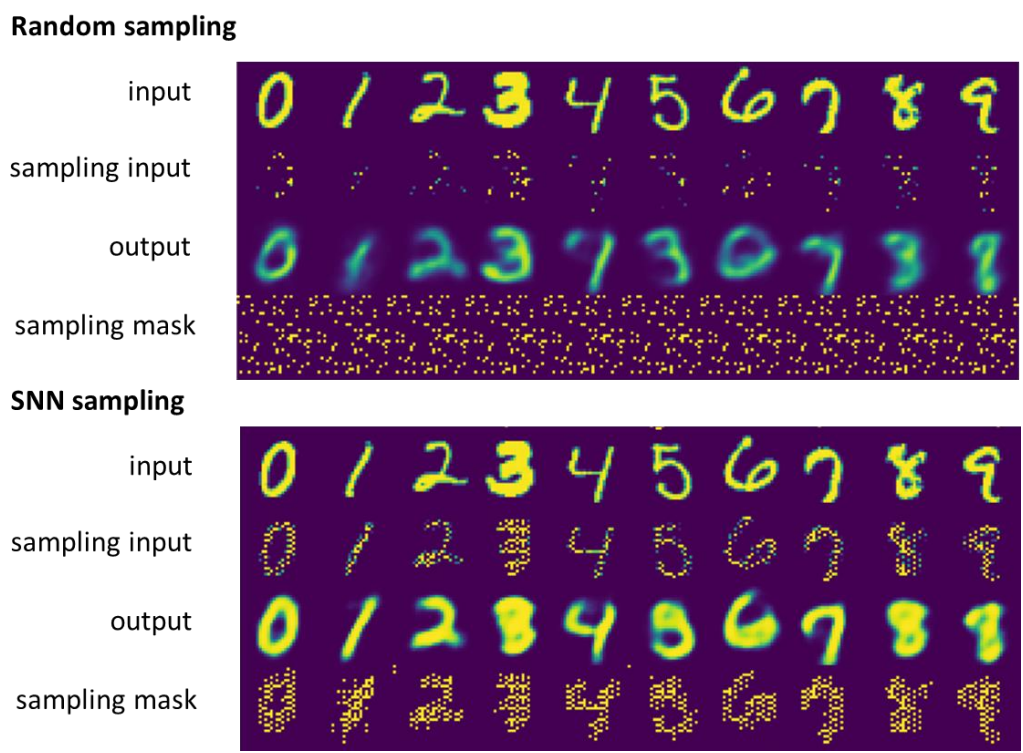

Fig. 6 Reconstruction comparison of 10% random sampling and 10% spiking sampling on the main reconstruction network trained by random sampling.

**Random sampling**

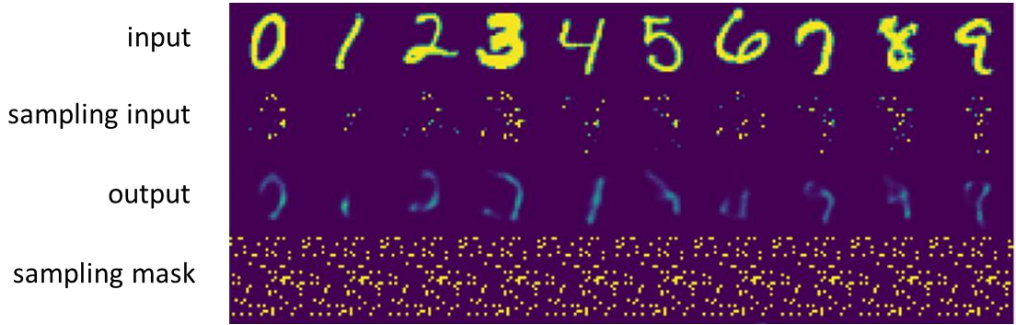

**SNN sampling**

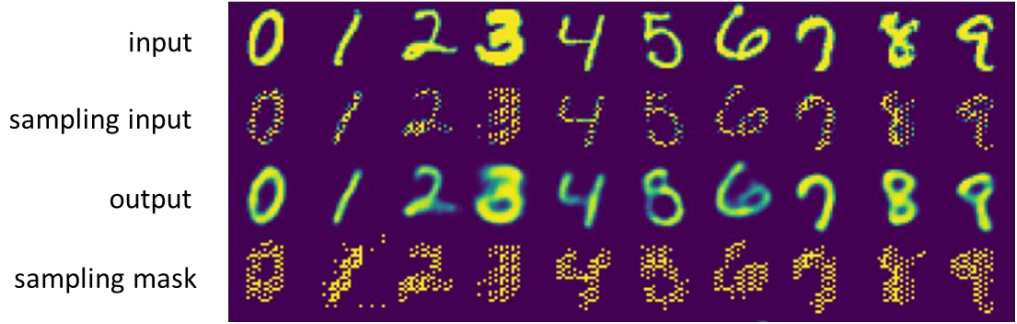

Fig. 7 Reconstruction comparison of 10% random sampling and 10% spiking sampling on the main reconstruction network trained by spiking sampling.

## 4. Conclusions

In this paper, we propose a novel sparse representation method by a spiking sampling neural network. We verify on static datasets that the network is able to learn sparse features of each sample independently by training. Compared to random sampling, the spiking sampling network performs better in image reconstruction. Our method can be applied to compress dynamic datasets with large amounts of data, which can greatly reduce the storage space and speed up data transfer.

## 5. Appendix

Network and training details. Table 1 shows network structures for image reconstruction on MNIST and CIFAR-10 and data compression on N-MNIST. Models are trained with MSE loss and Adam optimizer. The initial learning rate is set to 1e-4. SNN is trained by surrogate gradient [30]. The simulation time of SNN is 3 steps. For reconstruction, we trained models 100 epochs; for classification, we trained the model 20 epochs.

Table 1 Network structures for image reconstruction on
MNIST and CIFAR-10 and data compression on N-MNIST

| Dataset | Network Structure |
|---|---|
| MNIST | SNN: 16C3P1-MP2-4C3P1-MP2-16CT2S2-1CT2S2 |
| | Main network: FC784-FC256-FC64-FC20-FC64-FC256-FC784 |
| CIFAR-10 | SNN: 16C3P1-MP2-4C3P1-MP2-16CT2S2-1CT2S2 |
| | Main network: 12C4S2P1-24C4S2P1-48C4S2P1-96C4S2P1-48CT4S2P1-24CT4S2P1-12CT4S2P1-3CT4S2P1 |
| N-MNIST | SNN*:  12C4S2-24C4S2P1-48C4S2P1-96C4S2P1-48CT4S2P1-24CT4S2P1-12CT4S2P1-2CT4S2 |
| | Main network*: 64C3P1-64C3P1-64C3P1-64C3P1-2C3P1 |
| | Classification network: 128C3-128C3-MP2-FC2048-FC100-FC10 |

Note: nCm—Convolutional layer with n output channels, kernel size = m and stride = 1, nCm—transposed convolutional layer with n output channels, kernel size = m and stride = 1, MP2—2D max-pooling layer with kernel size = 2 and stride = 2, FC—FC layer. * represents all convolutional layers are 3D layers.

## 6.  Declaration

**Conflict of interest**  The authors declare that they have no known competing financial interests or personal relations that could have appeared to influence the work reported in this paper.

**Data availability**  The datasets generated during and/or analysed during the current study are available from the corresponding author on reasonable request.